\documentclass{article}
\usepackage{here}
\PassOptionsToPackage{numbers, compress}{natbib}

\usepackage[final]{neurips_2023}

\usepackage{graphicx}
\usepackage[utf8]{inputenc} 
\usepackage[T1]{fontenc}    
\usepackage{hyperref}       
\usepackage{url}            
\usepackage{booktabs}       
\usepackage{amsfonts}       
\usepackage{nicefrac}       
\usepackage{microtype}      
\usepackage{xcolor}         
\usepackage{ascmac}
\usepackage{fancybox}
\usepackage{listings,jvlisting} 
\title{
    JMedLoRA:Medical Domain Adaptation on Japanese Large Language Models using Instruction-tuning
}

\author{
  Issey Sukeda$^1$ \and Masahiro Suzuki$^2$ \and Hiroki Sakaji$^2$ \and Satoshi Kodera$^1$\\ 
  \\
  \\
  $^1$ Department of Cardiovascular Medicine, Graduate School of Medicine,\\
  $^2$ Department of Systems Innovation, School of Engineering,\\
  The University of Tokyo, Japan\\
}

\begin{document}
\maketitle
%

\begin{abstract}
    In the ongoing wave of impact driven by large language models (LLMs) like ChatGPT, the adaptation of LLMs to medical domain has emerged as a crucial research frontier.
    Since mainstream LLMs tend to be designed for general-purpose applications, constructing a medical LLM through domain adaptation is a huge challenge. 
    While instruction-tuning is used to fine-tune some LLMs, its precise roles in domain adaptation remain unknown. 
    Here we show the contribution of LoRA-based instruction-tuning to performance in Japanese medical question-answering tasks. 
    In doing so, we employ a multifaceted evaluation for multiple-choice questions, including scoring based on ``Exact match'' and ``Gestalt distance'' in addition to the conventional accuracy.
   Our findings suggest that LoRA-based instruction-tuning can partially incorporate domain-specific knowledge into LLMs, with larger models demonstrating more pronounced effects.
   Furthermore, our results underscore the potential of adapting English-centric models for Japanese applications in domain adaptation, while also highlighting the persisting limitations of Japanese-centric models.
   This initiative represents a pioneering effort in enabling medical institutions to fine-tune and operate models without relying on external services.
\end{abstract}

\section{Introduction}\label{sec:intro}
The study and development of medical large language models (LLMs)  like ChatGPT have the potential to revolutionize the field of medicine and healthcare in profound ways. These models, when fine-tuned and adapted to the medical domain, can assist healthcare professionals in numerous critical tasks, such as disease diagnosis, treatment planning, and patient care. Due to their vast language comprehension capabilities, LLMs can possibly provide up-to-date information, suggest evidence-based treatment options, and even predict disease outcomes with a high degree of accuracy.

To fill the gap between mainstream of LLMs and the practical use in clinical environments, domain adaptation is a crucial technique even after the surge in ChatGPT\footnote{\url{https://chat.openai.com/}}, a powerful LLM service that has revolutionized the way we interact with text and language by its astonishing ability to generate sentences. While these general-purpose models are powerful in zero-shot inference in unseen tasks, fine-tuned models may have the potential to outperform them in domain-specific tasks. Several instances of domain adaptation within the medical field in the context of the English-centric LLMs~\cite{singhal2023large,singhal2023towards,tu2023towards} exist, but research in this direction is largely lacking in Japanese, highlighting the need to pioneer studies in non-English contexts (It is worth noting that JMedRoBERTa~\cite{sugimoto_nlp2023_jmedroberta} deals with BERT-based models, not GPT).
The potential impact of large-scale medical LLMs provided in one's native language could be shared not only in Japan but also in other non-English-speaking countries.

Moreover, using ChatGPT is impeded in clinical practices due to the concerns related to data privacy and security. The potential risks associated with data breaches or misuse of confidential patient information underscore the need for robust security measures and ethical considerations, further complicating its seamless integration into clinical settings. Hence, we need to consider domain adaptation using other LLMs for incorporating medical knowledges. 
Recently, several parameter efficient fine-tuning methods have been proposed including Low Rank Adaptation (LoRA)~\cite{hu2021lora} and its quantized version (QLoRA)~\cite{dettmers2023qlora}, where only the limited parameters are chosen as the target of the fine-tuning. Performed along with instruction-tuning, LoRA has demonstrated some success in acquiring conversational abilities~\cite{suzuki2023base}. 
That being said, the ability and limitation of LoRA-based instruction-tuning has not been clarified in domain adaptation. Recently, ``Superficial Alignment Hypotheses'' has been proposed, insisting that fine-tuning does not contribute significantly to the acquisition of knowledge, but this topic remains controversial~\cite{zhou2023lima}. Therefore, we aim to investigate whether LoRA-based instruction tuning can be effective in acquiring domain-specific knowledge. 

The primary research questions guiding our study are as follows:

1. How and how much can domain knowledge be incorporated into LLMs by LoRA-based fine-tuning?

2. Do larger English-centric LLMs outperform smaller Japanese-centric LLMs?

3. Does the amount of fine-tuning hold significance?

To answer these questions, we conduct a comprehensive comparison between different LLMs fine-tuned with our own Japanese medical dataset. Each model is evaluated by medical question-answering tasks. In doing so, we clarify the ability and limitation of incorporating domain-specific knowledge by LoRA so as to open the path to further constructing stronger versions of various domain-specific Japanese LLMs.


\section{Methods}\label{sec:method}

We conduct a comprehensive comparison between different LLMs fine-tuned with Japanese medical dataset, including those we have created ourselves. 
The models used in our experiments are available at \url{https://huggingface.co/AIgroup-CVM-utokyohospital}.

    \subsection{Base Models}
    As our main scope is developing Japanese LLMs, we use as the baseline OpenCALM-7B\footnote{\url{https://huggingface.co/cyberagent/open-calm-7b}} , the open-sourced Japanese foundation LLM with 6.5 billion parameteres published by CyberAgent, Inc. 
    Additionally, we train a new model MedCALM, which is based on OpenCALM-7B and additionally pretrained on our own medical text dataset. The training dataset consists of 2420 examples, and the evaluation dataset has 50 examples. The maximum token count is set to 768, and the batch size is set to 63. The model was trained for 2000 steps.
    We further use Llama2-70B-chat-hf~\cite{touvron2023llama}\footnote{\url{https://huggingface.co/meta-Llama/Llama-2-70b-chat-hf}}, a powerful English-centric LLM released by Meta Inc.

    \subsection{Instruction-tuning}
    Instruction-tuning~\cite{wei2021finetuned} refers to the process of fine-tuning or optimizing the behavior and output of the model by providing explicit instructions or guidance as a prompt text during the generation of text. Recently, several implementations of instruction-tuning have been developed for GPT-based LLMs. Low Rank Adaptation (LoRA)~\cite{hu2021lora} is one of the major parameter-efficient fine-tuning methods provided in PEFT library~\cite{peft}. The quantized version of LoRA is provided as QLoRA~\cite{dettmers2023qlora}, which enables us to tune a model with as many as 70 billion parameters within limited computation resources. 
    In this study, we apply LoRA to OpenCALM-7B and QLoRA to Llama2-70B-chat-hf, respectively.

\section{Experiment}\label{sec:exp}
    We applied each fine-tuning method to each base model and verified accuracy and medical correctness of the generated responses. All experiments were conducted on NVIDIA A100 with 4 GPUs (80GB VRAM each).
    
    \subsection{Training}
    To perform instruction-tuning, we constructed a medical question-answer dataset containing 77422 records in instruction format by utilising ChatGPT (gpt-3.5-turbo). Two medical articles, one from the official journal of The Japanese Circulation Society and another from the Journal of the Japanese Society of Internal Medicine (JJSIM), were used as input for ChatGPT.
    For further details of instruction data generation, see Appendix \ref{appendix:template}. 
    The number of epochs and steps were set so as to approximately align with the overall computational time.

    \subsection{Evaluation by Medical Question-answering Tasks}
    To assess performance in medical domain, 
    we prepared two medical Q\&A datasets, IgakuQA dataset~\cite{jpn-med-exam_gpt4} and JJSIMQA. JJSIMQA is a dataset comprising 5-choice questions in JJSIM (see Appendix \ref{appendix:medicalQA}). 
    We evaluated the output responses of each model by three metrics. \textit{Gestalt score} is defined as the average of Gestalt distance between the response and the correct answer, which is calculated by a string matching algorithm that is based on the longest common subsequence.
    \textit{Accuracy} denotes the correctness rate by considering the choice closest to the model's response when measured using Gestalt distance as the model's output.
    Finally, \textit{Exact match} is defined as the proportion of responses that contain the correct answer.
    All evaluation metrics above take the value between 0 and 1, and the larger the better.

    All results are summarized in Table~\ref{table:results-med2-5}, where the performance by plain ChatGPT (gpt-3.5-turbo) is added to the last column for reference. Top 2 scores in each row are highlighted in bold.

    {\tabcolsep=3.6pt
    \begin{table}[tb]
      \caption{Performance of Japanese medical question-answering tasks. 0s and 1s denote 0-shot inference and 1-shot inference, respectively. The top 2 scores of each row are highlighted in bold. 0 step denotes the original base model.}
      \label{table:results-med2-5}
      \centering
      \begin{tabular}{|c|cccc|cccc|ccc|c|}\hline
        & \multicolumn{4}{|c|}{OpenCALM-7B}&\multicolumn{4}{|c|}{MedCALM} & \multicolumn{3}{|c|}{Llama2-70B} & ChatGPT \\ \hline
        Steps of (Q)LoRA & 0 & 1k &3k&10k& 0 & 1k &  3k &10k & 0 & 0.9k & 3k & - \\  \hline
        Accuracy (0s)& .222&.219&.213&.239&.236&.216&.214&.223&.216&\textbf{.295}&.285&\textbf{.438}\\
        Exact match (0s)&.002&.019&\textbf{.047}&.039&.028&.029&.029&.034&.017&.038&.045&\textbf{.112}\\
        Gestalt score (0s)&.033&.114&.141&.078&.032&.096&.113&.085&.071&.276&\textbf{.287}&\textbf{.369}\\
        Accuracy (1s)& .227&.244&.196&.223&.207&.188&.188&.204&.244&\textbf{.296}&\textbf{.267}&-\\
        Exact match (1s)& .009&\textbf{.070}&.026&.019&.021&0&.001&.019&.045&\textbf{.057}&.056&-\\
        Gestalt score (1s)&.053&.186&.087&.078&.028&0&.002&.035&.247&\textbf{.331}&\textbf{.314}&-\\
        Training hours&-&4.6&24.0&37.0&-&8.9&23.7&58.4&-&12.7&42.4&-\\
        \hline
      \end{tabular}
    \end{table}
    }
    
\section{Discussion}\label{sec:discussion}
    \subsection{Numerical Evaluation of the Effects of Fine-tuning}

    We have observed notable score improvements with LoRA after an appropriate number of steps, particularly with Llama2-70B showing the most significant enhancement. This suggests that utilizing a more powerful English-centric model as the base model holds promise for domain adaptation even in Japanese contexts.
    
    It has been controversial in instruct-tuning whether we should repeat epochs or just once. Our results show that a single epoch (1k steps) of instruction-tuning improves the performance but increasing the number of epochs exacerbates the model.
    Furthermore, additional pretraining did not contribute to performance improvement. Therefore, we conclude that conducting LoRA-based instruction-tuning for a single epoch without considering additional pretraining is a more practical and promising approach, especially when dealing with limited training data.
    
    Note that in this study, we exclusively utilized medical documents closely related to the task for additional pretraining. However, we believe that the efficacy of additional pretraining could be further explored by incorporating a broader range of medical domain documents or by extracting and expanding from a general-purpose corpus.
    Determining the necessary amount of data for additional pretraining to improve performance in downstream tasks is a challenge we will face in the future.
    
    \subsection{Deterioration of 1-shot Performance}
    From Table~\ref{table:results-med2-5}, it is evident that every OpenCALM-based model except the original one experiences a decline in 1-shot inference scores when compared to 0-shot. This outcome highlights the fact that the original OpenCALM model clearly loses its capability to leverage example responses provided within the context, whereas Llama2-70B retains this ability even after instruction-tuning.

    \subsection{Comparison on Evaluation Metrics}
    The evaluation of LLMs have been intensively argued recently. 
    Regarding the evaluation method of LLMs, there is still no unified ``rule-of-thumb'' method yet. 
    While the existing metrics (e.g. JGLUE~\cite{kurihara2022jglue}) or leaderboards (e.g. Nejumi LLM leaderboard~\footnote{\url{http://wandb.me/nejumi}}) can assess the fluency of generated texts, they do not adequately evaluate the accuracy of domain-specific knowledge. We wish to point out that three metrics used in our experiments also exhibit certain shortcomings. For example, the use of multiple-choices question for evaluating LLMs has been controversial~\cite{pezeshkpour2023large,zheng2023large}.
    \textit{Exact match} cannot accurately score responses that, while conveying the correct meaning, do not match the text verbatim. 
    \textit{Gestalt score} is asymmetric and prone to multiple choices. The development of even more superior evaluation metrics is eagerly anticipated.
    
    \subsection{Difficulty and Limitations}\label{subsec:limit}
        Not to mention that numerous LLM training techniques are still in the developmental stage, training medical LLMs presents several limitations. First and foremost, the quantity and quality of data could be insufficient in our work. Preparing a medical dataset in instructional format can be expensive. In this study, we employed ChatGPT for automated generation, but this approach may become financially burdensome when preparing larger datasets. Moreover, data cleansing has consistently posed challenges, and achieving perfect results in this work may not have been feasible.
        

        Moreover, during the writing period of this paper, Japanese LLMs that are considered to perform better than OpenCALM, which was used in this study, have been released (see e.g. Rakuda benchmark\footnote{\url{https://yuzuai.jp/benchmark}}). There is a possibility of obtaining different results when using them as the base model. 
%
\section{Conclusion and Future Work}\label{sec:conclusion}
    In this paper, we explore the capabilities and limitations of LoRA through various comparative analyses in medical domain. 
    LoRA-based instruction-tuning, while avoiding an excessive number of steps, can partially integrate domain-specific knowledge into LLMs, with larger models demonstrating more pronounced effects.
    We also observe  a decrease in performance by additional pretraining under scarce training dataset.
    Furthermore, our results underscore the potential of adapting larger English-centric models for Japanese applications in domain adaptation, while also highlighting the persisting limitations of Japanese-centric models including the deterioration of 1-shot performance after instruction-tuning. 
    Our findings here suggest that, at present, the most promising approach in constructing a domain-specific LLM is applying QLoRA to larger English-centric base models.
    
    That being said, the improvement in performance is not sufficient regarding the clinical implementations. To fully harness the potential of medical LLMs in healthcare settings, addressing both the performance limitations and the associated security and privacy concerns is imperative. Further research and development efforts are needed to enhance the accuracy and reliability of these models, ensuring they meet the rigorous standards required for clinical decision.


\bibliographystyle{plain}
\bibliography{neurips}

\begin{thebibliography}{10}

\bibitem{dettmers2023qlora}
Tim Dettmers, Artidoro Pagnoni, Ari Holtzman, and Luke Zettlemoyer.
\newblock Qlora: Efficient finetuning of quantized llms.
\newblock {\em arXiv e-prints}, pages arXiv--2305, 2023.

\bibitem{hu2021lora}
Edward~J Hu, Phillip Wallis, Zeyuan Allen-Zhu, Yuanzhi Li, Shean Wang, Lu~Wang, Weizhu Chen, et~al.
\newblock Lora: Low-rank adaptation of large language models.
\newblock In {\em International Conference on Learning Representations}, 2021.

\bibitem{jpn-med-exam_gpt4}
Jungo Kasai, Yuhei Kasai, Keisuke Sakaguchi, Yutaro Yamada, and Dragomir Radev.
\newblock Evaluating {GPT}-4 and {ChatGPT} on {J}apanese medical licensing examinations.
\newblock {\em arXiv preprint arXiv:2303.18027}, 2023.

\bibitem{kurihara2022jglue}
Kentaro Kurihara, Daisuke Kawahara, and Tomohide Shibata.
\newblock Jglue: Japanese general language understanding evaluation.
\newblock In {\em Proceedings of the Thirteenth Language Resources and Evaluation Conference}, pages 2957--2966, 2022.

\bibitem{peft}
Sourab Mangrulkar, Sylvain Gugger, Lysandre Debut, Younes Belkada, and Sayak Paul.
\newblock Peft: State-of-the-art parameter-efficient fine-tuning methods.
\newblock \url{https://github.com/huggingface/peft}, 2022.

\bibitem{pezeshkpour2023large}
Pouya Pezeshkpour and Estevam Hruschka.
\newblock Large language models sensitivity to the order of options in multiple-choice questions.
\newblock {\em arXiv preprint arXiv:2308.11483}, 2023.

\bibitem{singhal2023large}
Karan Singhal, Shekoofeh Azizi, Tao Tu, S~Sara Mahdavi, Jason Wei, Hyung~Won Chung, Nathan Scales, Ajay Tanwani, Heather Cole-Lewis, Stephen Pfohl, et~al.
\newblock Large language models encode clinical knowledge.
\newblock {\em Nature}, pages 1--9, 2023.

\bibitem{singhal2023towards}
Karan Singhal, Tao Tu, Juraj Gottweis, Rory Sayres, Ellery Wulczyn, Le~Hou, Kevin Clark, Stephen Pfohl, Heather Cole-Lewis, Darlene Neal, et~al.
\newblock Towards expert-level medical question answering with large language models.
\newblock {\em arXiv preprint arXiv:2305.09617}, 2023.

\bibitem{sugimoto_nlp2023_jmedroberta}
Kaito Sugimoto, Taichi Iki, Yuki Chida, Teruhito Kanazawa, and Akiko Aizawa.
\newblock J{M}ed{R}o{BERT}a: a japanese pre-trained language model on academic articles in medical sciences (in {J}apanese).
\newblock In {\em Proceedings of the 29th Annual Meeting of the Association for Natural Language Processing}, 2023.

\bibitem{suzuki2023base}
Masahiro Suzuki, Masanori Hirano, and Hiroki Sakaji.
\newblock {From Base to Conversational: Japanese Instruction Dataset and Tuning Large Language Models}.
\newblock {\em arXiv preprint arXiv:2309.03412}, 2023.

\bibitem{touvron2023llama}
Hugo Touvron, Louis Martin, Kevin Stone, Peter Albert, Amjad Almahairi, Yasmine Babaei, Nikolay Bashlykov, Soumya Batra, Prajjwal Bhargava, Shruti Bhosale, et~al.
\newblock Llama 2: Open foundation and fine-tuned chat models.
\newblock {\em arXiv preprint arXiv:2307.09288}, 2023.

\bibitem{tu2023towards}
Tao Tu, Shekoofeh Azizi, Danny Driess, Mike Schaekermann, Mohamed Amin, Pi-Chuan Chang, Andrew Carroll, Chuck Lau, Ryutaro Tanno, Ira Ktena, et~al.
\newblock Towards generalist biomedical ai.
\newblock {\em arXiv preprint arXiv:2307.14334}, 2023.

\bibitem{wei2021finetuned}
Jason Wei, Maarten Bosma, Vincent Zhao, Kelvin Guu, Adams~Wei Yu, Brian Lester, Nan Du, Andrew~M Dai, and Quoc~V Le.
\newblock Finetuned language models are zero-shot learners.
\newblock In {\em International Conference on Learning Representations}, 2022.

\bibitem{zheng2023large}
Chujie Zheng, Hao Zhou, Fandong Meng, Jie Zhou, and Minlie Huang.
\newblock On large language models' selection bias in multi-choice questions.
\newblock {\em arXiv preprint arXiv:2309.03882}, 2023.

\bibitem{zhou2023lima}
Chunting Zhou, Pengfei Liu, Puxin Xu, Srini Iyer, Jiao Sun, Yuning Mao, Xuezhe Ma, Avia Efrat, Ping Yu, Lili Yu, et~al.
\newblock Lima: Less is more for alignment.
\newblock {\em arXiv preprint arXiv:2305.11206}, 2023.

\end{thebibliography}

\newpage
\appendix

\section{Medical Q\&A dataset}\label{appendix:medicalQA}

For performance evaluation, we prepare two medical question-answer dataset, IgakuQA and JJSIMQA.






\begin{itembox}[l]{An example from IgakuQA (originally in Japanese)}
"problem\_id": "116A1",

"problem\_text": "Which of the following is incorrect regarding hypertension caused by obstructive sleep apnea?",

"choices": \{"a":"It often leads to nocturnal hypertension.", "b":"Weight reduction is recommended for obese patients.", "c":"Alpha-blockers are the first-line choice of medication.", "d":"Morning hypertension is frequently observed in home blood pressure measurements.", "e":"Continuous positive airway pressure (CPAP) therapy is expected to lower blood pressure."\}, 

"text\_only": true, 

"answer": ["c"], 
\end{itembox}





\begin{itembox}[l]{An example from JJSIMQA(5-choice questions in JJSIM)  (originally in Japanese)}
"problem\_text": "Which of the following is incorrect about recent cases of hepatitis B in Japan? Choose one.",

"choices":\{"a":"The HBs antigen positivity rate has significantly decreased due to the initiation of mother-to-child infection prevention programs.", "b":"HBV (hepatitis B virus) genotype Ae can become a carrier through horizontal transmission in adults.", "c":"In Japan, routine HBV vaccination began in October 2016.", "d":"HBV genotype C is more prevalent in the Tohoku and Miyako-Yaeyama regions.", "e":"Horizontal transmission of HBV during childhood is thought to be partly attributed to father-to-child transmission and communal living."\}, 

"text\_only": true, 

"answer": ["d"], 
\end{itembox}

\section{Sample questions and responses}\label{appendix:sample}

We randomly create questions that ask to each model the treatment to a symptom. Table~\ref{tab:hikakuhyou} shows the responses of each model to the given input ``$\#\#\#$ Instruction: Please provide detailed instructions for the treatment to be administered to patients with the following diseases. $\#\#\#$ Input: deep vein thrombosis''\footnote{Originally in Japanese.}.

\begin{table}    
\caption{Sample responses to the input ``$\#\#\#$ Instruction: Please provide detailed instructions for the treatment to be administered to patients with the following diseases. $\#\#\#$ Input: deep vein thrombosis'' }
  \label{tab:hikakuhyou}
  \centering
  \begin{tabular}{|c|c|l|}\hline
    model&LoRA/QLoRA&response (originally in Japanese)\\ \hline \hline
    OpenCALM&none&\begin{tabular}{l}
At the bedside, the nurse encouraged the patient to move \\ their legs to assist in getting out of bed. The patient attempted to \\flex and extend their legs; however, they were unable to move \\the leg joints and needed assistance to stand up independently. \\ The nurse approached the patient sitting on a bedside chair, \\provided shoulder support, lifted their arms, extended their \\elbows, and helped them rise slowly. \\The nurse then gently released their grip and returned to the chair.\\ The nurse proceeded to call the next patient.
\end{tabular} \\ \hline
    OpenCALM& 1 epoch of LoRA&\begin{tabular}{l}The procedure to be carried out for patients with the following\\ disease involves the implementation of appropriate treatment\\ and preventive measures
    \end{tabular} \\ \hline
    OpenCALM&10 epoch of LoRA& \begin{tabular}{l}Detailed treatment methods have not been provided.    \end{tabular}\\ \hline
    Llama2-70B-chat-hf&none& \begin{tabular}{l}In some cases, a procedure called thrombol \footnote{Originally in English.} \end{tabular} \\ \hline
    Llama2-70B-chat-hf&900 steps of QLoRA&\begin{tabular}{l}The treatment of deep vein thrombosis involves the use of\\ thrombolytic agents and anticoagulants.\end{tabular}\\ \hline
    Llama2-70B-chat-hf&3000 steps of QLoRA& \begin{tabular}{l}The treatment for deep vein thrombosis includes \\pharmacological therapy to dissolve the blood clot\\ as well as surgical interventions to remove the thrombus.\end{tabular}\\ \hline
  \end{tabular}
\end{table}

\section{Instruction data generation by ChatGPT}\label{appendix:template}
The prompt we use to create instruction dataset is the following.

\begin{itembox}[l]{Prompt to generate instruction dataset by ChatGPT\footnote{Originally written in Japanese.}}
\#\#\# Instructions:\\
You are a machine designed to generate various question and answer pairs. Please create data with question (instruction) and answer (output) pairs based on the following input, considering it as prior knowledge. Format the data as {'instruction': question content, 'output': answer content}, and do not include line breaks. Repeat this process 15 times and list one data pair per line.\\
\#\#\# Input: \\
\{input\_text\}
\end{itembox}


\section{Hyperparameters}

\begin{table}    
\caption{LoRA parameters}
  \label{table:lora}
  \centering
  \begin{tabular}{|c|c|c|}\hline
    base model&OpenCALM-7B&Llama2-70B\\ \hline
    fine-tuning method&LoRA&QLoRA\\
    learning rate &5e-5&2e-4\\
    input length&512&512\\
    target max length&512&512\\
    batch size&8&8\\
    fine-tuning steps&1k, 3k, 10k& 0.9k, 3k\\
    $r$ of (Q)LoRA&8&64\\
    $\alpha$ of (Q)LoRA &32&16\\
    dropout rate of (Q)LoRA&0.05&0.1\\
    target parameter&query\_key\_value&all linear layers\\
    \hline
  \end{tabular}
\end{table}

Parameters in fine-tuning steps are listed in Table~\ref{table:lora}. All follows the default parameter setting of PEFT library and qlora library, respectively.

For the text generation part, temperature is set to 0.1, max\_new\_tokens is set to 256, top\_p is set to 0.9, and repetition\_penalty is set to 1.05. 

\section{Prompt Templates}
In order to let models answer the given questions by text generation function, we provided the following prompt texts, Japanese one for OpenCALM-based models and English one for Llama2-based models. Both of them have the same meaning in each language. 

\begin{figure}[tb]
        \centering
            
             
             
             
             

            
             
             
             
             

             
             
             
             
             

        
        
        

        \includegraphics[keepaspectratio, scale=0.5]{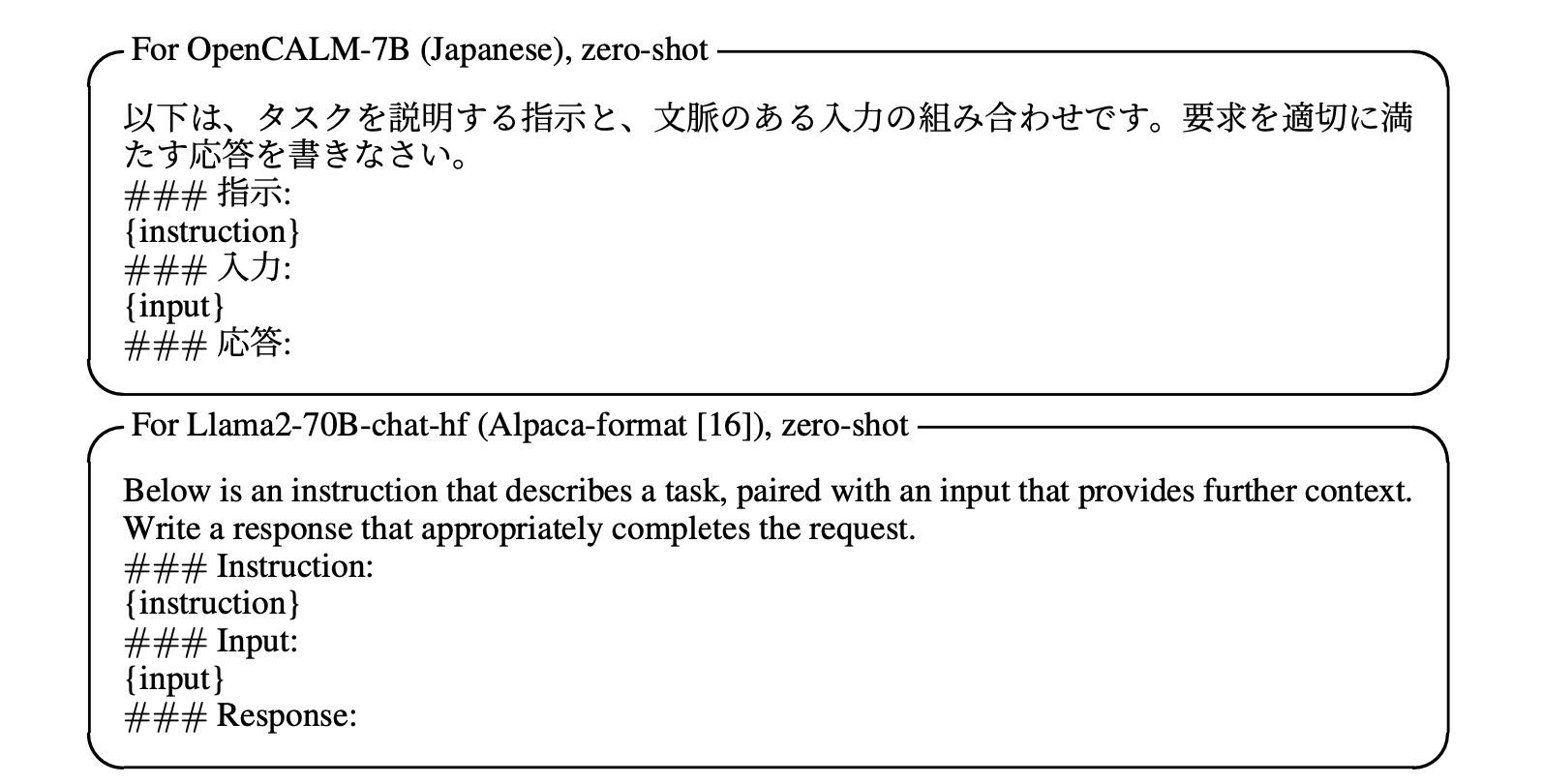}
        \caption{
           Prompt templates used in our instruction-tuning
        }
        \label{fig:template}
    \end{figure}

For each sample from Medical Q\&A dataset, "problem\_text" is incorporated in \{instruction\} and "choices" is incorporated in \{input\} in Figure~\ref{fig:template}.

\end{document}